\def\year{2018}\relax
\documentclass[letterpaper]{article} %DO NOT CHANGE THIS
\usepackage{aaai18}  %Required
\usepackage{times}  %Required
\usepackage{helvet}  %Required
\usepackage{courier}  %Required 
\usepackage{url}  %Required
\usepackage{graphicx}  %Required
\usepackage{xcolor}

\title{The IQ of Neural Networks}
\author{Dokhyam Hoshen \and Michael Werman \\School of Computer Science and Engineering
\\The Hebrew University of Jerusalem
\\ Jerusalem, Israel}
 
\begin{document}
% The file aaai.sty is the style file for AAAI Press 
% proceedings, working notes, and technical reports.
%

\maketitle
\begin{abstract}
% \begin{quote}
IQ tests are an accepted method for assessing human intelligence. The tests consist of several parts that must be solved under a time constraint. Of all the tested abilities, pattern recognition has been found to have the highest correlation with general intelligence. This is primarily because pattern recognition is the ability to find order in a noisy environment, a necessary skill for intelligent agents. In this paper, we propose a convolutional neural network (CNN) model for solving geometric pattern recognition problems. The CNN receives as input multiple ordered input images and outputs the next image according to the pattern. Our CNN is able to solve problems involving rotation, reflection, color, size and shape patterns and score within the top 5\% of human performance.
% \end{quote}
\end{abstract}

\noindent Evaluating machine learning methods against human performance has served as a useful benchmark for artificial intelligence. Deep learning methods have recently achieved super-human level performance on several important tasks such as face recognition \cite{taigman2014deepface} and Large Vocabulary Continuous Speech Recognition (LVCSR) \cite{xiong2017microsoft}. Although the above tasks might have some correlation with human intelligence, they are not testing intelligence directly. In this paper, we take a more direct route and evaluate neural networks directly on tasks designed to test human intelligence. 

Intelligence Quotient (IQ) tests are one of the most common methods of defining and testing human computation and comprehension abilities \cite{stern1914psychological}. IQ tests are only estimates of intelligence (as opposed to actual measurements) and their effectiveness is in active debate \cite{flynn1987massive,siegel1989iq}. Despite the controversy, IQ tests are used in practice to make
critical decisions, such as school and workplace placement, and assessment of mental disabilities. IQ scores have also been shown to correlate with morbidity and mortality, socioeconomic background and parental IQ.  

IQ tests measure different skills: verbal intelligence, mathematical abilities, spatial reasoning, classification skills, logical reasoning and pattern recognition. In this paper, we focus on the pattern recognition section. In these problems, the test-taker is presented with a series of shapes with geometric transformations between them. As in most IQ tests, the problem is presented with multiple choices for answer. One of these answers is the (most) correct one. A limited number of  possible transformations  are used to produce the questions, thus it is possible to gain experience and learn  to solve them easily and quickly. Nevertheless, pattern recognition problems have added benefits in comparison to the other types of questions, as they rely less on a standard educational background (as in the verbal or mathematical) and therefore (arguably) are less biased towards upbringing and formal education. Pattern recognition problems are also related to  real life problems and are therefore assumed to have a high correlation with the actual intelligence.

In this paper, we solve IQ-test style pattern recognition questions with a deep network. We train our network on computer generated
pattern recognition problems. The input to each example consists of two images, where the second image is a transformed version of the first image. The models infer the transformation and output the prediction of the next image in two different forms (evaluated separately):
\begin{enumerate}
\item Multiple choice: The model is presented with a  number of possible  images and choses the most probable one.  
\item Regression: The model draws the most likely next image.
\end{enumerate} 
% Both methods were not given any prior knowledge of the specific functions of the transformations . 
Simply by showing the model sequences of images we were able to grasp geometric concepts such as shape and rotation.

We trained the network on series of images exhibiting the following transformations (see Fig.~\ref{fig:data_build} for examples):
\begin{itemize}
\item Rotation: Each shape is rotated by a constant angle with respect to the previous shape.
\item Size: Each shape is expanded or shrunk by a fixed scaling factor with respect to the previous shape.
\item Reflection: Each shape is transformed by a combination of a rotation (as before) and a reflection.
\item Number: Each image contains one more polygon than in the previous one i.e if the image contains two triangles the second and third images will contain three and four triangles, respectively. 
\item Color: The model is presented with a series of objects in different colors and must decide the color of the next object. 
\item Addition: The model is presented with a series of shapes that have another shape added to them. The model must decide what a  combination of these previous images  looks like.

\end{itemize}

We performed extensive experiments investigating the performance of CNNs at IQ tests. We trained the networks on a dataset containing all the  transformation options. In this experiment, the operations were also used simultaneously, i.e there can be two shapes in the image: one rotating and one expanding. In a different set of experiments, for each operation, we trained a specialized network on a dataset containing only the specific operation, for analysis purposes. 

This research brings us a step closer to comparing deep learning intelligence to human intelligence. It is also important for evaluating the difficulty of IQ tasks. We can also use this model in adapting tests to the level of the individual and so improve the current methods of evaluation. Additionally, probabilities give an indication to how confusing different choices were, thus allowing a more accurate evaluation.

\section{Related Work}
\label{sec:related}

% Solving IQ tests has not been widely researched in computer
% science and machine learning. Nevertheless, 

Psychologists
have tried to estimate the effectiveness of different questions
in intelligence tests \cite{crawford2001nart}. A commonly
used and well validated test for pattern recognition assessment
is  Raven’s Progressive Matrices, the test taker is typically presented with 60 non-verbal multiple choice questions (as shown in Figure 1). Such tests are popular mainly due to their independence from language and writing skills.

\begin{figure}
  \includegraphics[width=0.2\textwidth]{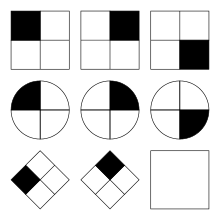}
  \caption{ Example of Raven’s test style questions. Given
eight shapes the subject must identify the missing piece.}
\label{fig:example}
\end{figure}

One of the advantages of visual pattern recognition problems in IQ tests is that they requires a minimal amount of prior knowledge. Therefore, such tests can also be used to quantify the intelligence of very young children who have not yet learned basic mathematical and verbal concepts. This measure is thus less polluted by environmental factors. Work was done on the relationship between the scores that children obtained in these problems in infancy and their results in later years, in IQ tests in general as well as specifically the pattern recognition section \cite{dilalla1990infant}. The score was found to be a good predictor for later years, but adding   birth order and parental education resulted in  improved predictions. 

The subject of solving Raven's Progressive Matrices problems computationally was studied by \cite{kunda2009addressing}. The focus of this work was finding a feature representation for solving Raven's Progressive Matrices with simple classifiers. While their research was the first to address these problems as a computational problem, they did not publish enough experimental results to validate their method. Our approach is different and more general, we use deep neural networks to automatically learn the representation jointly with the classifier, rather than handcrafting the feature representation.

Work was done by \cite{wang2015solving} on automatic solution of the verbal reasoning part of IQ tests. Machine learning methods using hand-crafted features were able to automatically solve verbal reasoning questions featuring synonyms and antonyms as well as word analogies. This line of work is related to our research as it solves "analogy" problems, where the subject needs to grasp the transformation rules between words and generalize them. It
deals linguistic transformations rather than 
visual ones.

\cite{hoshen2016visual} analyzed the capability of DNNs  learning arithmetic operations. In this work, the network learned the concept of addition of numbers based on end-to-end visual learning. This shows the possibility of learning arithmetic transformations without prior knowledge of basic concepts such as "number" or "addition".

Much research has been done on Frame Prediction, predicting the next frame given one or more frames.  Mathieu et al. \cite{mathieu2015deep} presented a video next frame prediction network. Due to uncertainty is the output of real video, a GAN loss function was used and shown to improve over the standard euclidean loss. A CNN system was presented by Oh et al. for frame prediction in Atari games \cite{oh2015action}. This is simpler than natural video due to both having simpler images and more deterministic transformations. Some work  been done on object and action prediction from video frames by \cite{vondrick2016anticipating}, in which the agent learned to predict the FC7 features and used it to classify future actions. In this paper, we deal with data that uses pre-specified transformations and simple shapes and is, therefore, more amenable to detailed analysis than previous research.

Our architecture uses Convolutional Neural Networks (CNNs) \cite{lecun1990handwritten}.
For multiple choice type problems, we use a classification architecture containing  a discriminator (see e.g. \cite{krizhevsky2012imagenet}). For open questions, in order to generate an image we use an autoencoder type architecture (see e.g. \cite{hinton2006reducing}). The hyper-parameters of our CNN architecture, are inspired by the architecture used in DCGAN \cite{radford2015unsupervised}. We optimized the network parameters using backpropogation \cite{rumelhart1988learning} \cite{lecun2012efficient}, with the ADAM \cite{kingma2014adam} gradient descent algorithm.

\begin{figure*}[t]
\centering
\includegraphics[width=0.9\textwidth]{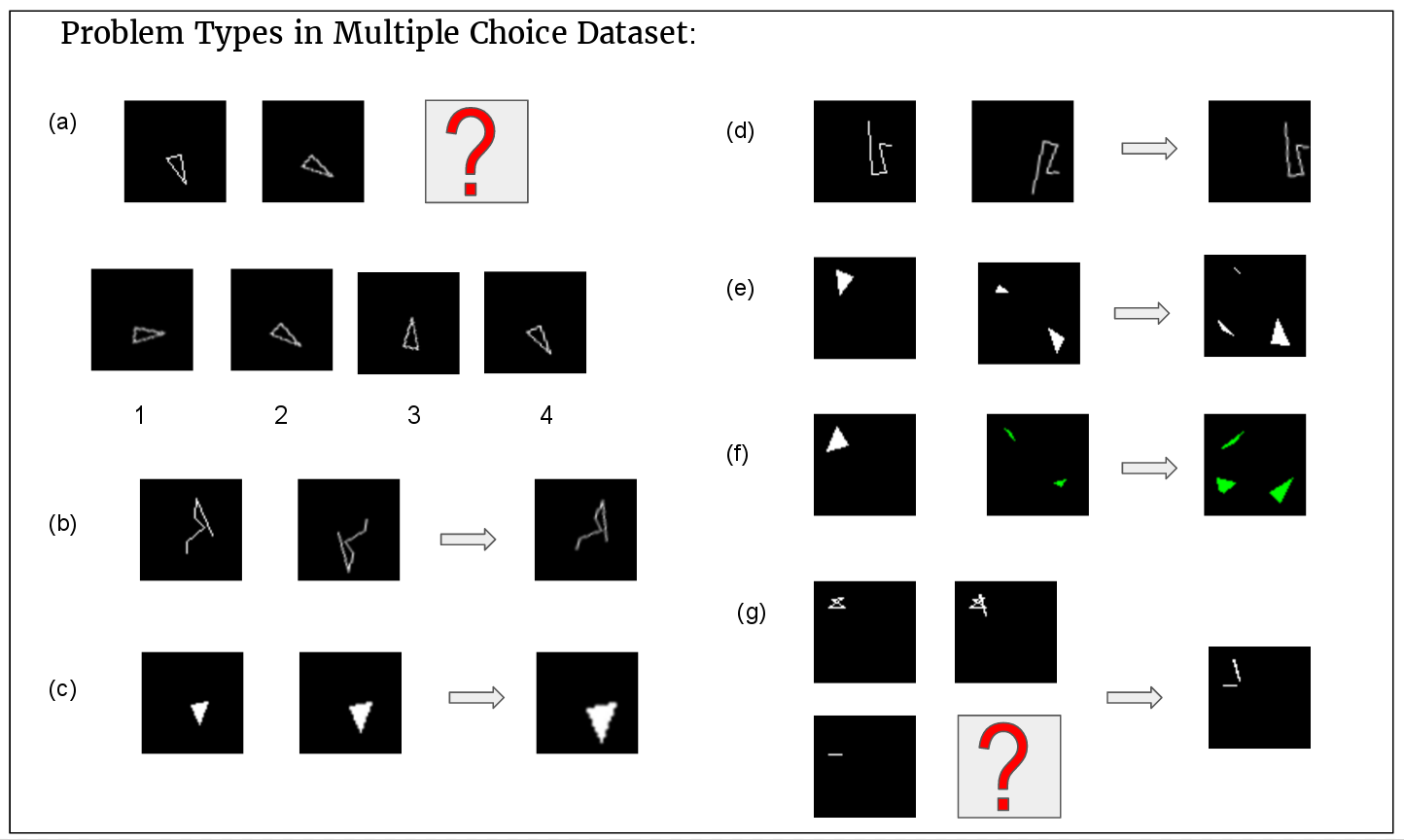}
  \caption{\label{fig:data_build} Multiple choice question problem types. In this figure each example corresponds to a different problem type. For each problem, there are 4 possible answers and only 1 is correct (shown in full only for (a), for space considerations).  (a) Polygon rotation: A polygon is rotated by a constant angle $\theta$, between consecutive frames. The correct answer is: Option 1. (b) Squiggle Rotation: Six random  points are sampled and rotated by a constant angle, $\theta$. (c) Size: A polygon is dilated by a constant scale factor, $\mu$. (d) Reflection: A shape is rotated by a constant angle  $\theta$ and then reflected around the $y$-axis. (e) Counting: Each frame shows one more polygon than in the previous one. (f) Color: The third frame shows one more polygon than the second frame in the color of the second image (green).(g) Image Addition: The second image adds a line to the first image and the fourth image adds that same line to the third image.}
\end{figure*}

\section{Creating the Dataset}

In this paper, we evaluate two types of IQ test scenarios: solving multiple choice pattern recognition problems and producing the next image.

\textbf{\textit{Multiple choice questions:}} This scenario is used in most IQ tests. The model receives six input images: 2 input images and 4 options for the next image. At train time, it receives an index corresponding to the correct answer. The model outputs the predicted probability for each of the presented options.  The most probable option is picked as the model's answer.

We form the questions in the following way: for each sample question, we randomly choose a transformation $T$ from the following: rotation, size, reflection, number, color and addition. We then randomly produce a 64x64 image containing a shape $S$. The shape is randomly selected out of  triangles, squares, circles or random squiggles. We apply the chosen transformation on the first image resulting in the second frame $TS$. We produce the correct answer, $TTS$, and the 3 wrong answers (the other options) using other transformations. 
\begin{itemize}
\item Rotation: we randomly select an angle $\theta \in [0..2\pi]$ and rotate the shape by $\theta$. The wrong answers are created by sampling different angles for rotation or a different operation. See Fig.~\ref{fig:data_build}(a)(b).

\begin{figure*}[t]
  \includegraphics[scale=0.3]{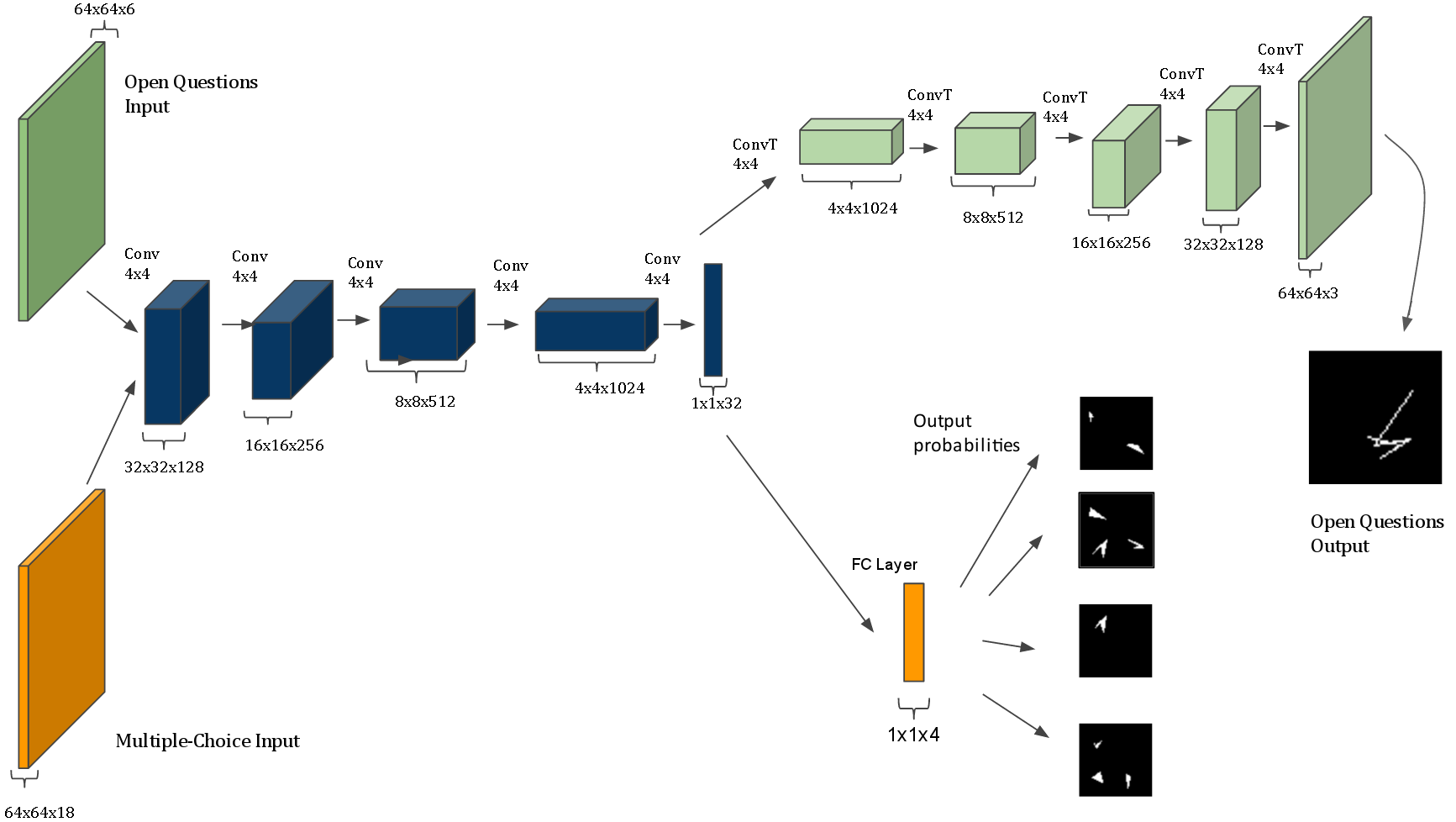}
  \caption{Model architectures: In this figure, we show the architectures of both models, the open questions and the multiple choice questions. In blue we have the CNN that is used by both models, the encoder network. The inputs of the models are different: the multiple answer questions require four extra images  from which to choose the answer. Thus, the input for the multiple-choice problems consists of 6 images while the input for the open questions consists of 2 images. After the code generation, the models split: the autoencoder  runs the decoder  (green) and outputs an image. The CNN classifier (used for the multiple-choice problems) has one more affine layer (orange) and outputs four probabilities, one for each  option. Note that the two models are trained separately and do not share features. }
  \label{fig:both_arch}
\end{figure*}

\item Size: We randomly select a scale parameter $\mu \in [0.5..2]$. The wrong answers are created by sampling a different resizing ratio or a different operation. See Fig.~\ref{fig:data_build}(c).

\item Reflection: we randomly select an angle for rotation and rotate the shape after reflecting it at each stage. The wrong answers are created using different operations. See Fig.~\ref{fig:data_build}(d).

\item Number: Between 1 and  4 shapes are selected. For each successive image we add one further exemplar of the shape (this shape has the same number of vertices as the previous shape but will most likely not be a replica of the previous shape) i.e if the first frame contains a triangle, the next image will contain two triangles. The wrong images will show a different number of shapes than the correct one or it will show the correct number of shapes but of the wrong shape. See Fig.~\ref{fig:data_build}(e).

\item Color: Two colors and a shape are selected at random. The first image is the shape with the first color and the second with two shapes with the second color. The solution should be the three copies of the shape with the second color. In the example in Fig. \ref{fig:data_build}(f) the correct image has three green triangles. The wrong answers will include different colors, shapes and transformations. This is in essence a combination of the "number" questions with color elements.
\item Addition:  Three shapes are randomly selected.The first image contains one of these shapes.The second contains  both the first and second shapes. The third  contains just the third image. 
The correct answer is the union of images two and three but not including the part of image 2 that was inherited from image 1. Formally, $im4 = im3 \cup im2 \setminus im1 $. This problem is a combination of image addition and subtraction. See Fig.~\ref{fig:data_build}(g).

\end{itemize}

\textbf{\textit{Open questions:}} In this scenario, the model is required to  predict the next frame. At training time we do not provide the model with four options to choose from, as the model generates the next frame on its own. Since it is difficult to evaluate the correctness of the predicted frame for the "number" transformation, we did not include it in the open questions dataset. Furthermore, the "color" transformation in the this scenario is defined differently. For each sample two colors and shapes are selected at random. The first image  shows the first shape in the first color, the second  shows the second shape in the second color. The third image (the solution) should show the first shape in the second color. 

We automatically generated 100K images for training and 1000 images for testing, according to the rules described above. The data was normalized to have a normal mean and standard deviation at each pixel.

We will publicly release the dataset generation Python code.

\section{Model Architecture}
 For multiple choice problems we used a standard CNN and for open questions, we used an autoencoder. Both models were implemented in PyTorch.

\textbf{\textit{Multiple choice questions:}} We used a CNN architecture with 4 convolutional layers, each followed by batch normalization and ReLU activation layers. We found this (shallow) depth to work the best. The input layer receives six images, the first two frames of the sequence and the four options that the model must choose the most likely answer from. The output layer is a linear layer with Softmax activation, it outputs the probability of each of the 4 optional images to be the correct answer. The model architecture is sketched in Fig.~\ref{fig:both_arch}. We can additionally use the probabilities to infer which questions were challenging for the model (e.g. by the probability vector entropy).

\textbf{\textit{Open questions:}} In this scenario, we generated the answer image rather than simply selecting from a set of option images. This required  a network architecture containing both an encoder and a decoder (an autoencoder). The choice of hyper-parameters used in our architecture borrows heavily from the one used in DC-GAN \cite{radford2015unsupervised}. The encoder consists of 5 convolutional layers, each with 4x4 convolution kernels with a stride of 2 (no pooling layers, this is implicitly done by the convolution operation's stride). Each convolution is followed by leakyReLU activation layers with a leakage coefficient of 0.2. The input of the decoder is the first two RGB input  frames (total size: 64x64x6). The output of the encoder is a code vector $z$ of dimension 32. The decoder receives the code $z$ as input and outputs a full image (size: 64x64x3). The decoder is comprised of 5 deconvolution layers (Transposed Convolution layers) with 4x4 convolution kernels and stride 2, this effectively upsamples by a factor of 2 with each deconvolution layer. Each deconvolution layer is followed by an ELU activation layer. See Figure \ref{fig:both_arch} for a diagram of our architecture.

For both scenarios, the same optimization settings were used. We optimized the model using Adam \cite{kingma2014adam}. The learning rate was annealed by a factor of $0.1$ every 100 epochs. We trained the model using mini-batches of size 64. The loss is calculated by the cross-entropy loss. 

The number of training epochs required for convergence was different for each problem as was their convergence rate. The number of epochs to convergence ranged between 30 and 100 depending on the problem.

\section{Results}\label{sec:Results}
In this paper, we set to investigate if neural networks are able to solve geometric pattern recognition problem as used in IQ tests. In this section, we evaluate the performance of our CNN models on learning the transformations to the geometric shapes (rotation, size, reflection etc). For both models, the results convincingly show that the networks have grasped the basic concepts.

We present  quantitative evaluations of model performance. 

\textbf{\textit{Multiple-answer questions:}} 
As this is a classification problem, a quantitative evaluation is simple for this scenario. The accuracy rates of our model on the different tasks are presented in Tab.~\ref{tab:mainres}.  We can see that the overall score was significantly smaller than most of the individual scores. This is most likely due to error of classification of the transformation, a problem that does not exist in individual trainings. Out of all the transformations, the one with the highest error rate was the $number$ transformation. E.g. one failure case had two similar options, with squares (wrong answer) having two vertices that were illegibly close to one another that looking very similar to triangles (correct answer).

\begin{figure}[t]
  \includegraphics[width=0.47\textwidth]{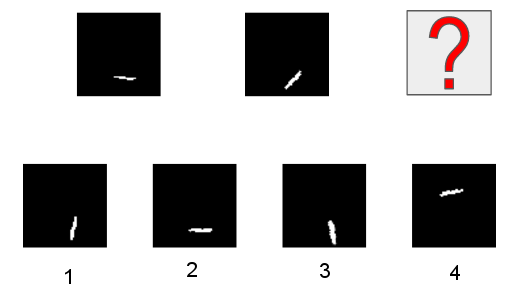}   
  \caption{Wrong Prediction on a rotation problem. True answer: 3 Prediction: 2 
  Probabilities: Option 1 $\approx$ 0. Option 2 $\approx$ 0.9. Option 3 $\approx$ 0.09. Option 4 $\approx$ 0. We can see that the network was confused by the thin shape and had difficulty deciding which direction the shape was rotating.}
  \label{fig:wrong_ex}
\end{figure}

In Fig.~\ref{fig:wrong_ex} we show an example of the questions where the network classified the wrong option as the correct answer. We added the probabilities that the model assigned to these options as well since this is indicative of the network's error rate. 

It visually apparent that in this failure case, the options were very close to one another, and that the network's output probabilities were also very close for these options (however the wrong option had a slightly higher probability). This indicates that the failure does not signify the network's failure to learn these transformations and the actual error rate is smaller.
\begin{table}[t]
\centering
  \begin{tabular}{|l|c|r|}
      \hline
      \textbf{Transformation} & \textbf{Success rate} \\
      \hline

      Rotation Polygon & 98.4\% \\
      \hline
      Rotation Squiggle &  94\% \\
      \hline
      Size & 100\%\\
      \hline
       Reflection & 100\% \\
      \hline
      Color & 95.6\% \\
      \hline
      Number & 91.4\%\\
      \hline
      Addition & 100\%\\
      \hline
      All-Transformation Network & \textbf{92.8\%}  \\
      \hline

  \end{tabular}
 \caption{ Accuracies for multiple-choice questions: We can see that the all-transformation network accuracy was lower than most of the individual scores. This is most likely due to additional errors in classification of the transformation, a problem that does not exist in individual trainings. Out of all the transformations, the the $number$ transformation had the highest error rate.}
 \label{tab:mainres}
\end{table}

\begin{figure*}
\centering
  \includegraphics[height=\textheight, width=0.9\textwidth]{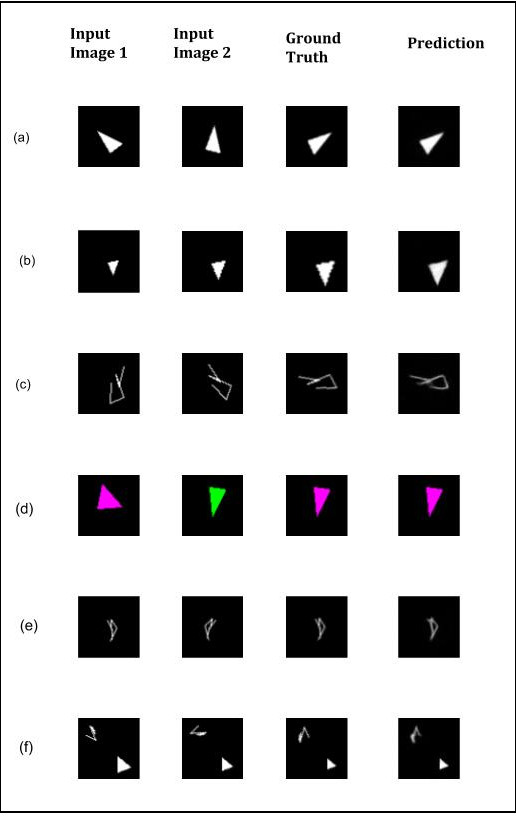}
  \caption{ Open questions Scenario. We present output examples for the following tasks: (a) Rotation (b) Size (c) "Squiggle" Rotation (d) Color (e) Reflection (f) Combination of Rotation + Size}
  \label{fig:all_res}
\end{figure*}

\textbf{\textit{Open questions:}} We trained and tested the autoencoder network on the following problems: rotation, size, color, and reflection. We also tested the capabilities of the agent on tasks where the transformation was a combination of two of the basic transformations. Some questions included two shapes, each simultaneously changing according to a different transformation. In Fig. \ref{fig:all_res} we show examples from the different types of questions that were used for evaluation. 

It is apparent that while the network has managed to grasp the concepts of the transformations in all cases, there is some blurring as the finer details in the image are lost in the encoding (downsampling) stage of the autoencoder. We measure prediction accuracy by mean squared error (MSE) between the ground truth and predicted image.

 In Tab.~\ref{tab:mse} we present the average MSEs for each problem type.
\begin{table}[t]
\centering
\begin{tabular}{|l|c|r|}
      \hline
      \textbf{Transformation} & \textbf{Average MSE} \\
      \hline
      Rotation Polygon &  4.1E-4 \\
      \hline
      Rotation Squiggle &  5.7E-4\\
      \hline
      Size &  2.4E-4 \\
      \hline
       Reflection & 4.1E-4  \\
      \hline
      Color & 2.1E-4\\
      \hline
      \textbf{Overall} & \textbf{3.5E-4}  \\
      \hline
  \end{tabular}
 \caption{ The results show that the reflection and rotation transformations are more difficult to learn. Empirical tests also show that these transformations are the most difficult for humans to solve. }
 \label{tab:mse}
\end{table}
We can see from the results that reflection and rotation tasks were the hardest to learn (as shown by the relatively high error). A possible theoretical explanation for the relative difficulty of learning the rotation transformation is that more parameters need to be inferred to define the rotation matrix: the center point coordinates and the angle. As our reflection problems also include a rotation, the same number of parameters much be inferred. In comparison, for color problems only a single parameter must be inferred, as it leaves one image unchanged but multiplies its values by an identical constant that multiplied the other image. Only one parameter is also needed for the scale factor in the size questions. Inferring more parameters can result in a more challenging task. Furthermore, empirical research has found that rotation questions are significantly more difficult for humans to solve in IQ tests \cite{jaarsveld2010solving}. 

\subsection{Performance on Noisy Images}
As pattern recognition is often required in noisy environments, we tested the model's performance on noisy images. The setup is identical to the noiseless dataset, but with IID Gaussian noise of $\sigma=99$ added to all input and output images. 

The model performed well in the noisy setting. In the multiple-choice scenario, the average model error rate was $87\%$ compared to $91\%$ in the noiseless case. In the open question scenario, the result images were generated correctly without the added noise, as noise cannot be predicted. 

\subsection{Performance on Real Questions}
One of the main objectives of this paper is to compare neural network performance on benchmarks used to evaluate human intelligence. When constructing the dataset, we analyzed IQ tests used in practice in the aim of building a realistic dataset. The rules used for dataset construction are similar as those used for creating human IQ tests. 

We tested our model on a test set of 40 questions taken from IQ tests used by the National Institute for Testing and Evaluation in Israel. The questions mostly consisted of reflection and addition problems, which are prominent in human IQ tests. Our model scored 38 correct answers out of 40 (5\% error rate). This lies within the top 
$5\%$ of human performance for all ages in the range between 20 and 70 according to the Institute's measurements and similar performance thresholds are given by  \cite{raven2000raven}. This performance percentile corresponds to an IQ of above 127 \cite{hunt2010human}. The true IQ is potentially higher, as the test does not discriminate between the top IQ percentages. 

One difference between our dataset and IQ tests used in practice is the complexity of the shapes (more edges, more types of shapes etc.) and the number of options to choose from. While the number of options can easily be dealt with by small modification to the architectures, the complexity of the shapes  pose an obstacle for our model. It was not able to successfully generate crisp pictures in the Open Questions format. We believe that by making our dataset more picture-realistic we can improve transfer learning in this scenario. 

\begin{figure}
  \includegraphics[width=0.4\textwidth]{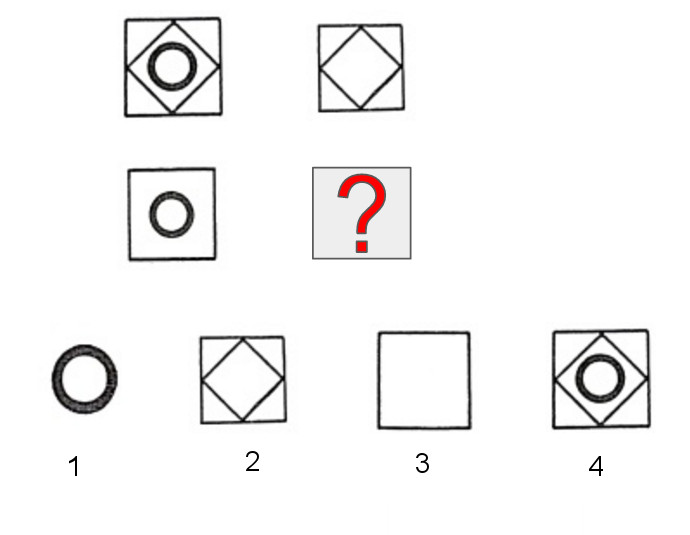}
  \caption{ Sample Question from the Test Set. Answer: 3}
  \label{fig:real_ex}
\end{figure}

\section{Conclusion and Future Work}
We  presented an effective neural network model for solving geometric pattern recognition problems as used in IQ tests. We also presented a model that is capable of generating the images of the answers, giving more insight into the patterns that our model is able to learn.

More research can be done on the types of questions that CNNs can solve, or more importantly, the types of problems that they fail at. In the future, we would like to explore more challenging patterns that capture the limit of CNN learning capacity. Subjects with high scores are typically tested on questions of "Advanced Progressive Matrices", but these questions were not compatible with the current architecture of our network. We would like to adjust our model to all types of questions that appear in Raven's Progressive Matrices test.
\newpage
\newpage
\bibliographystyle{aaai}
\bibliography{sample}

\end{document}